\documentclass[10pt,twocolumn,letterpaper]{article}
\usepackage{cvpr}              
\usepackage{multirow}
\usepackage{makecell}
\usepackage{appendix}

\definecolor{cvprblue}{rgb}{0.21,0.49,0.74}
\usepackage[pagebackref,breaklinks,colorlinks,allcolors=cvprblue]{hyperref}

\newcommand\nnfootnote[1]{
  \begin{NoHyper}
  \renewcommand\thefootnote{}\footnote{#1}
  \addtocounter{footnote}{-1}
  \end{NoHyper}
}

\title{SemanticGen: Video Generation in Semantic Space}

\author{\textbf{Jianhong Bai\textsuperscript{$\rm 1^{\ast}$}}, \textbf{Xiaoshi Wu\textsuperscript{$\rm 2^{\dagger}$}}, \textbf{Xintao Wang\textsuperscript{\rm 2}}, \textbf{Xiao Fu\textsuperscript{\rm 3}}, \textbf{Yuanxing Zhang\textsuperscript{\rm 2}}, \textbf{Qinghe Wang\textsuperscript{\rm 4}},\\\textbf{Xiaoyu Shi\textsuperscript{\rm 2}}, \textbf{Menghan Xia\textsuperscript{\rm 5}}, \textbf{Zuozhu Liu\textsuperscript{\rm 2}}, \textbf{Haoji Hu\textsuperscript{$\rm 1^{\dagger}$}}, \textbf{Pengfei Wan\textsuperscript{\rm 2}}, \textbf{Kun Gai\textsuperscript{\rm 2}} \\
\textsuperscript{\rm 1}Zhejiang University, \textsuperscript{\rm 2}Kling Team, Kuaishou Technology, \textsuperscript{\rm 3}CUHK, \textsuperscript{\rm 4}DLUT, \textsuperscript{\rm 5}HUST\\
}

\newcommand{\teaser}{
\centering
\vspace{-2em}
{Project webpage:} \url{https://jianhongbai.github.io/SemanticGen/}
\includegraphics[width=1\textwidth,trim=0em 0em 0em 0em,clip]{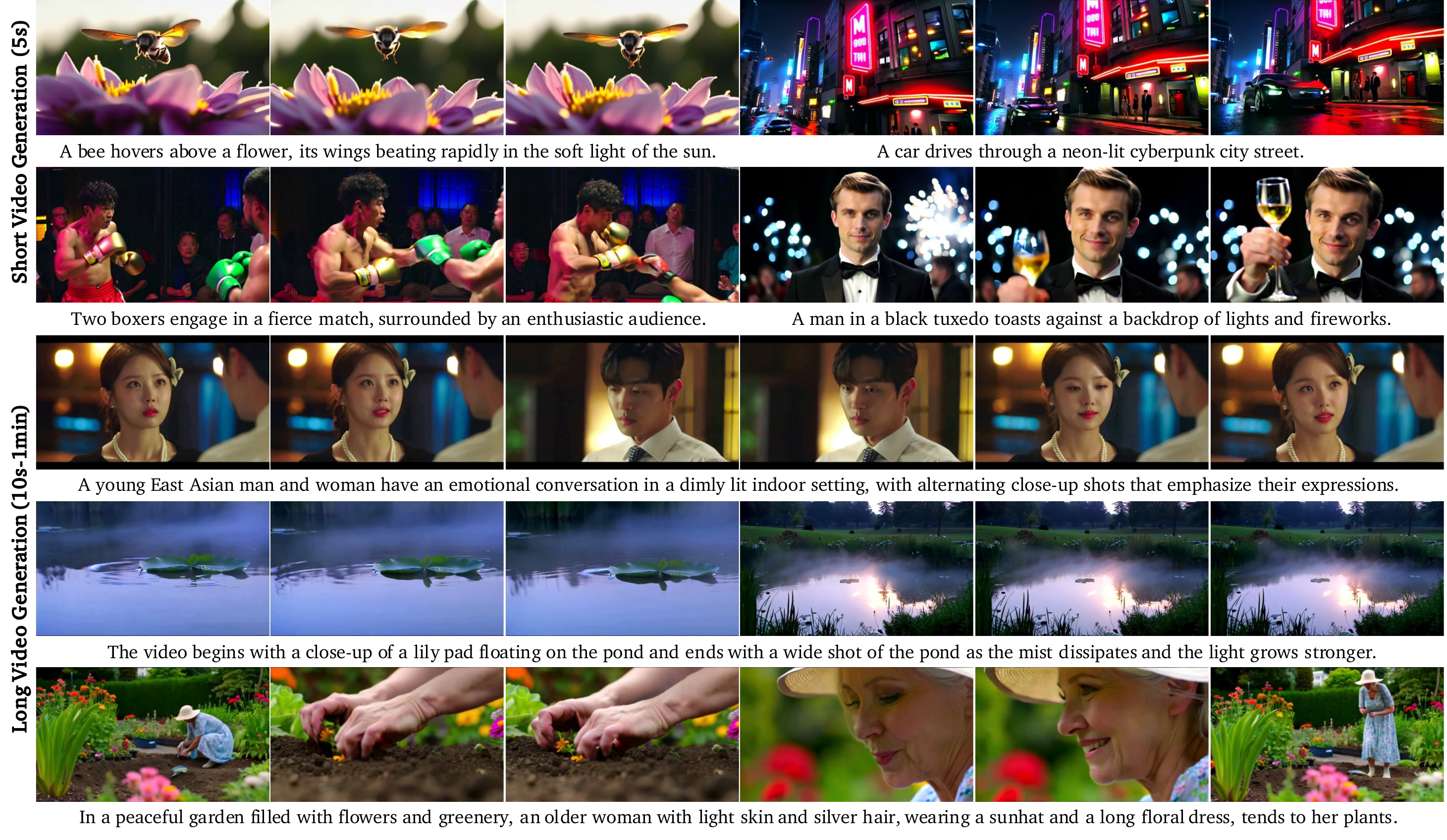}
\vspace{-2em}
\captionof{figure}{\textbf{Examples synthesized by SemanticGen.} SemanticGen generates high-quality videos from text prompts in the semantic representation space and scales to long-form generation of up to one-minute videos. Video results are on our \href{https://jianhongbai.github.io/SemanticGen/}{project page}.
}
\vspace{0.5em}
\label{fig_1}
}

\begin{document}
\twocolumn[
\maketitle
\teaser
]
\nnfootnote{$\ast$ Work done during an internship at Kling Team, Kuaishou Tech.} \nnfootnote{$\dagger$ Corresponding authors.}
\begin{abstract}

State-of-the-art video generative models typically learn the distribution of video latents in the VAE space and map them to pixels using a VAE decoder. While this approach can generate high-quality videos, it suffers from slow convergence and is computationally expensive when generating long videos.
In this paper, we introduce SemanticGen, a novel solution to address these limitations by generating videos in the semantic space. Our main insight is that, due to the inherent redundancy in videos, the generation process should begin in a compact, high-level semantic space for global planning, followed by the addition of high-frequency details, rather than directly modeling a vast set of low-level video tokens using bi-directional attention.
SemanticGen adopts a two-stage generation process. In the first stage, a diffusion model generates compact semantic video features, which define the global layout of the video. In the second stage, another diffusion model generates VAE latents conditioned on these semantic features to produce the final output.
We observe that generation in the semantic space leads to faster convergence compared to the VAE latent space. Our method is also effective and computationally efficient when extended to long video generation. Extensive experiments demonstrate that SemanticGen produces high-quality videos and outperforms state-of-the-art approaches and strong baselines.

\end{abstract}    
\section{Introduction}
\label{sec:intro}
Video generative models \cite{sora, veo3, wan} have made rapid advancements in recent years. Mainstream diffusion-based \cite{ddpm, ddim, flow-matching} methods first train a variational autoencoder (VAE) \cite{vae} with a reconstruction objective to project videos from pixel space into a latent space, and then train a diffusion model to fit the distribution of VAE latents. While effective, this paradigm has two key limitations. First, it suffers from slow convergence speed. To attain high-quality videos, existing methods often rely on extremely large computational budgets — on the order of hundreds of thousands of GPU-hours \cite{seawead} — highlighting the need for more compute-efficient training paradigms. Second, scaling to long videos remains challenging. Since modern VAEs typically have modest compression ratios, a 60s, 480p, 24fps video clip expands to over 0.5M tokens, making bidirectional full-attention diffusion modeling impractical. Although previous works attempt to reduce complexity via sparse attention \cite{xi2025sparse, xia2025training} or to adopt an autoregressive \cite{genie, videopoet} or diffusion–autoregressive hybrid \cite{diffusion-forcing, self-forcing} video generation framework, they often suffer from temporal drift or noticeable degradation in visual quality.

\begin{figure}[t]\centering
\includegraphics[width=0.49\textwidth]{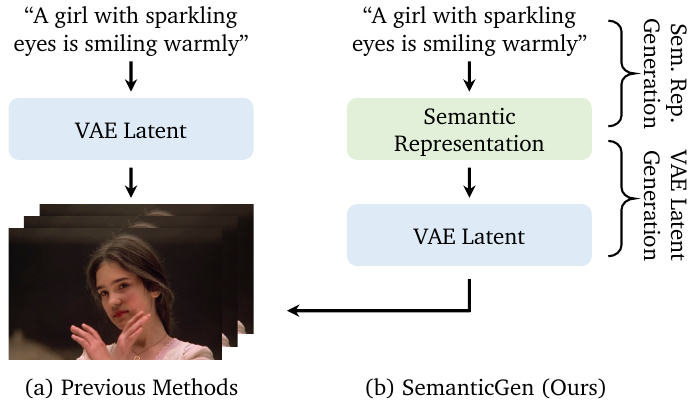}
\vspace{-0.5cm}
\caption{\textbf{Illustration of the proposed SemanticGen.}}
    \vspace{-0.5cm}
    \label{fig_semanticgen}
\end{figure}

To address these limitations, we propose SemanticGen, a framework that generates videos in a high-level semantic space before refining details in the VAE latent space, as illustrated in \cref{fig_semanticgen}. Our key insight is that, given the substantial redundancy inherent in videos, generation should first occur in a compact semantic space for global planning, and add high-frequency details afterwards — rather than directly modeling vast collections of low-level video tokens.
Technically, SemanticGen follows a two-stage paradigm. We first train a video diffusion model to denoise VAE latents conditioned on semantic representations, then train a semantic representation generator for high-level semantic modeling. We leverage off-the-shelf video-understanding tokenizers as the semantic encoder. However, we observe that directly sampling in high-dimensional semantic space could result in slower convergence speed and inferior performance. To this end, we propose semantic space compression via a lightweight MLP for effective training and sampling. Finally, we inject the compressed semantic embeddings into the video generator. Beyond improving training efficiency, this design extends naturally to long video generation, mitigating drift without sacrificing fidelity.

In addition to the proposed SemanticGen, recent studies extensively explore integrating semantic representations into video generation. A series of works \cite{vavae, dc-ae, maetok} incorporate semantic-level objectives into VAEs to learn semantic-rich and easily generable representations. These approaches are orthogonal to SemanticGen, as our framework is not tied to a specific VAE tokenizer. Another line of research optimizes latent generative models using semantic features. For example, REPA \cite{repa} aligns generative model hidden states with semantic representations to accelerate convergence, while RCG \cite{rcg} employs a two-stage process for unconditional image generation by first modeling semantic features and then mapping them to VAE latents. TokensGen \cite{tokensgen} is the most related method, as it also adopts a two-stage paradigm for video generation but further compresses VAE latents instead of using semantic features. However, we find that generating in the semantic space is fundamentally different from modeling in the compressed VAE space. In particular, the semantic space exhibits substantially faster convergence, as shown in \cref{fig_vae_vit}.

Experimental results demonstrate that our method outperforms state-of-the-art approaches and strong baselines in generating both short and long videos. Ablation studies further validate the effectiveness of our key design choices. The main contributions of this work are as follows:
\begin{itemize}
    \item We propose SemanticGen, a novel video generation framework that initially models in a compact semantic space before mapping to a low-level latent space.
    \item We identify key requirements for semantic encoders in video generation and develop semantic representation compression to effectively integrate semantic representations into the generation process.
    \item We conduct comprehensive experiments and ablation studies to demonstrate that SemanticGen provides significant advantages in terms of convergence speed and outperforms baselines in long video generation.
\end{itemize}
\section{Related Works}
\label{sec:related}
\subsection{Video Generative Models}

Recent advancements in video generative models can be roughly divided into diffusion-based approaches, autoregressive approaches, and their hybrid variants. Diffusion-based methods \cite{ddpm, ddim, flow-matching} model all frames with bidirectional attention and generate all frames simultaneously \cite{ho2022video, imagen, ltx-video, videocrafter2, wang2025cinemaster}. Early attempts \cite{animatediff, svd, lumiere, lavie, blattmann2023align} extended text-to-image models \cite{sd} with temporal modeling layers. Subsequent works \cite{wan, hunyuanvideo, mochi, cogvideox} benefit from the scalability of Diffusion Transformers \cite{dit}, achieving high-fidelity short video generation. However, due to the quadratic complexity of full sequence attention, its effective scalability to long video generation scenarios is limited. Autoregressive techniques \cite{genie, videopoet, weissenborn2019scaling, deng2024autoregressive, yin2025slow} generate each frame or patch of the video sequentially, facilitating applications like real-time video generation \cite{lin2025autoregressive, chen2025midas}, long-video generation \cite{wang2024loong, far}, etc. Meanwhile, a line of works \cite{jin2024pyramidal, hu2024acdit, diffusion-forcing, self-forcing, self-forcing++, zhou2025taming} adopts a diffusion-autoregressive hybrid paradigm, aiming to combine the advantages of both paradigms. Representative works, such as diffusion-forcing \cite{diffusion-forcing}, use a time-varying noise scheduler to achieve both causal modeling and full sequence attention. Self-forcing \cite{self-forcing} builds upon this and addresses the training-inference gap, reducing error accumulation issues. Nevertheless, these two types of methods generally exhibit inferior performance compared to diffusion-based methods. In this paper, we propose a novel diffusion-based video generation paradigm that achieves faster convergence and can effectively generalize to long video generation \cite{ttt, kim2024fifo, lct, framepack, longlive, holocine, wang2025multishotmaster}.
\subsection{Semantic Representation for Generation}

Recent studies demonstrate that incorporating semantic representations can significantly enhance the performance of generative models. One line of research \cite{vavae, dc-ae, unitok, rae, svg} focuses on introducing semantic representations to optimize the tokenizer of generative models. VA-VAE \cite{vavae} aligns VAE \cite{vae} latents with pre-trained semantic representations \cite{dinov2, mae}, while DC-AE \cite{dc-ae} and MAETok \cite{maetok} integrate semantic objectives \cite{mae} into VAE training. More recent work, RAE \cite{rae}, directly uses a semantic encoder in generative tasks and trains the corresponding decoder with a reconstruction objective. SVG \cite{svg}, building on self-supervised representations, additionally trains a residual encoder for improved reconstruction. These approaches consistently yield faster convergence and stronger image generation performance compared to using raw VAE latents.
Another line of methods directly optimizes latent generators \cite{rcg, repa, ddt, wu2025representation, kouzelis2025boosting}. RCG \cite{rcg} proposes first modeling self-supervised representations and then mapping them to the image distribution. REPA \cite{repa} aligns noisy input states in diffusion models with representations from pretrained semantic visual encoders. DDT \cite{ddt} uses decoupled diffusion transformers to separately learn semantic representations and high-frequency details. X-Omni \cite{x-omni} generates discrete semantic tokens with a unified autoregressive model and decodes with a diffusion generator. Our method fine-tunes diffusion models to learn compressed semantic representations that are subsequently mapped to the VAE latent space. This design leads to notably faster convergence than generating directly in the VAE latent space and effectively scales to long video generation.
\section{Method}
In this section, we present the design of SemanticGen. To generate the video clip $V$ from the text prompt, we leverage pre-trained video diffusion models (\cref{sec:preliminary}) and semantic encoders. We start by training a video diffusion model to generate video VAE latents conditioned on their compressed semantic representations (\cref{fig:pipe}a, \cref{sec:stage2}). Next, we learn the distribution of the compressed semantic representation from the text input (\cref{fig:pipe}b, \cref{sec:stage1}). During inference, we first generate the semantic representation, then map it to the VAE latent space (\cref{fig:pipe}c). We also demonstrate that our model can effectively generalize to long video generation (\cref{sec:longvideo}). 

\definecolor{lightgreen}{HTML}{E2F0D9}
\definecolor{lightyellow}{HTML}{FFF2CC}
\definecolor{lightblue}{HTML}{C4D8FB}
\definecolor{lightgray}{HTML}{E7E6E6}

\begin{figure*}[t]\centering
\vspace{-0.3cm}
\includegraphics[width=0.93\textwidth]{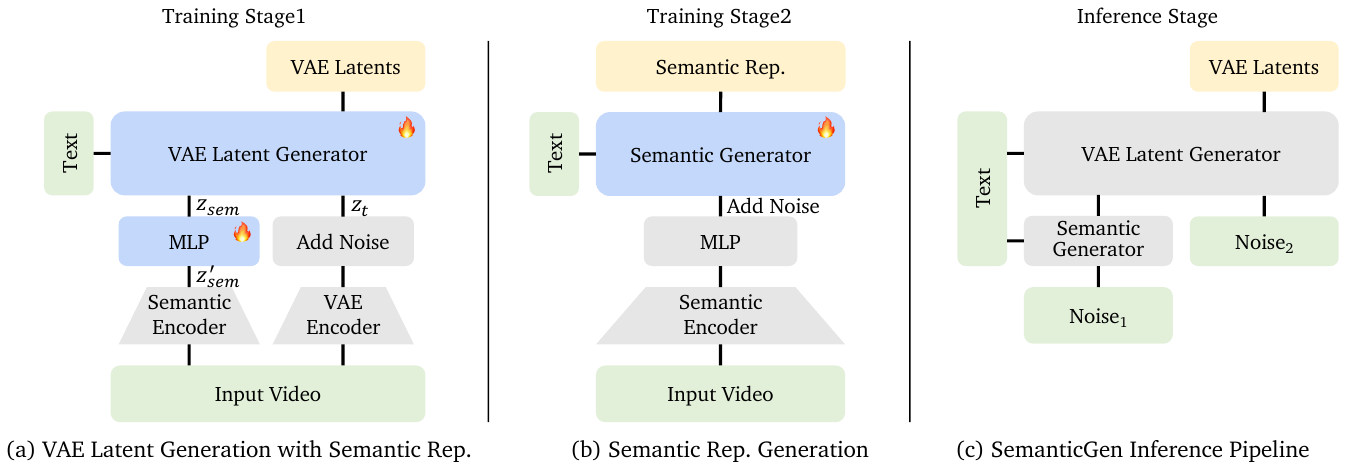}\vspace{-0.5em}
\caption{\textbf{Overview of SemanticGen.} (a) We optimize a latent diffusion model for denoising video VAE latents conditioned on their compressed semantic representations. (b) We train a semantic generator to fit the compressed semantic representation distribution of off-the-shelf semantic encoders. (c) During inference, we integrate the semantic generator and VAE latent generator to achieve high-quality T2V generation. \raisebox{0pt}[0pt][0pt]{\colorbox{lightgreen!60}{Green:}} Input; \raisebox{0pt}[0pt][0pt]{\colorbox{lightyellow!60}{Yellow:}} Output; \raisebox{0pt}[0pt][0pt]{\colorbox{lightblue!60}{Blue:}} Trainable models; \raisebox{0pt}[0pt][0pt]{\colorbox{lightgray!60}{Gray:}} Frozen models.}
    \label{fig:pipe}
\vspace{-1.0em}
\end{figure*}

\subsection{Preliminary: Text-to-Video Base Model}
\label{sec:preliminary}
Our study is conducted over an internal pre-trained text-to-video foundation model. It is a latent video diffusion model, consisting of a 3D Variational Autoencoder (VAE)~\citep{vae} and a Transformer-based diffusion model (DiT)~\citep{dit}. Typically, each Transformer block is instantiated as a sequence of spatial attention, 3D (spatial-temporal) attention, and cross-attention modules. The generative model adopts the Rectified Flow framework~\citep{esser2024scaling} for the noise schedule and denoising process. The forward process is defined as a straight path between the data distribution and a standard normal distribution, i.e.
\begin{equation}\label{eq:forward}
    z_t = (1-t)z_0 + t\epsilon,
\end{equation}
where $\epsilon \in \mathcal{N}(0,I)$ and $t$ denotes the iterative timestep.
To solve the denoising processing, we define a mapping between samples $z_1$ from a noise distribution $p_1$ to samples
$z_0$ from a data distribution $p_0$ in terms of an ordinary differential equation (ODE), namely:
\begin{equation}\label{eq:ODE}
dz_t=v_{\Theta}(z_t,t)dt, 
\end{equation}
where the velocity $v$ is parameterized by the weights $\Theta$ of a neural network. For training, we regress a vector field $u_t$ that generates a probability path between $p_0$ and $p_1$ via Conditional Flow Matching~\citep{flow-matching}:
\begin{equation}\label{eq:loss}
    \mathcal{L}_{LCM}=\mathbb{E}_{t,p_t(z,\epsilon),p(\epsilon)} ||v_{\Theta}(z_t,t)-u_t(z_0|\epsilon)||_2^2,
\end{equation}
where $u_t(z,\epsilon):=\psi_t^{'}(\psi_t^{-1}(z|\epsilon)|\epsilon)$ with {$\psi(\cdot|\epsilon)$} denotes the function of Eq.~\ref{eq:forward}.
For inference, we employ Euler discretization for Eq. \ref{eq:ODE} and perform discretization over the timestep interval at $[0, 1]$, starting at $t=1$. We then processed with iterative sampling with:
\begin{equation}\label{eq:euler_sample}
z_t=z_{t-1} + v_{\Theta}(z_{t-1},t) * \Delta t.
\end{equation}

\subsection{Video Generation with Semantic Embeddings}
\label{sec:stage2}
SemanticGen aims to generate videos by leveraging their compact semantic representations. Hence, we first fine-tune a pre-trained video diffusion model to learn denoising VAE latents conditioned on semantic representations. 
\paragraph{What Kinds of Semantic Encoders Are Needed for Video Generation?}
To select suitable off-the-shelf semantic encoders for video generation tasks, we identify three key requirements. \textbf{First,} the semantic tokenizer must be pre-trained on a large-scale \textit{video} dataset, as this enables the model to capture temporal semantics, such as object motion and camera movement. State-of-the-art image tokenizers, such as SigLip2 \cite{siglip2} and DINOv3 \cite{dinov3}, are trained solely on image datasets, and therefore are unable to effectively model temporal information. \textbf{Second,} the output representations should be compact in both spatial and temporal dimensions. The key insight is that, due to the inherent high redundancy in videos, generation should first occur in a high-level compact semantic space for global planning, followed by the addition of visual details. \textbf{Lastly}, the semantic tokenizer should be trained on a variety of video lengths, resolutions, thereby supporting the generation of diverse video content with variable length, aspect ratios, etc. 

As a result, we utilize the vision tower of Qwen-2.5-VL \cite{qwen25vl} as our semantic encoder, which is trained with a vision-language alignment objective \cite{clip} on image and video datasets. For video input, it first samples video frames at a lower fps (default 2.0), compresses 14x14 image patches into a single token, and then further compresses along each dimension by a factor of 2. This process ultimately transforms a video $V \in \mathbb{R}^{3 \times F \times H \times W}$, into a semantic representation $z_{sem}^{'} \in \mathbb{R}^{d \times F_s/2 \times H/28 \times W/28}$, where $d$ is the embedding dimension, and $F_s$ is the number of video frames sampled as input to the Qwen2.5-VL vision tower. 
Note that the framework proposed in this paper does not rely on using a specific semantic tokenizer. Other video semantic tokenizers, such as V-JEPA 2 \cite{v-jepa2}, VideoMAE 2 \cite{videomae2}, and 4DS \cite{4ds}, are also compatible with SemanticGen. In this paper, we use Qwen-2.5-VL to validate the effectiveness of SemanticGen, and we leave the systematic exploration of other semantic encoders as future work.
\begin{figure}[t]\centering
\includegraphics[width=0.48\textwidth]{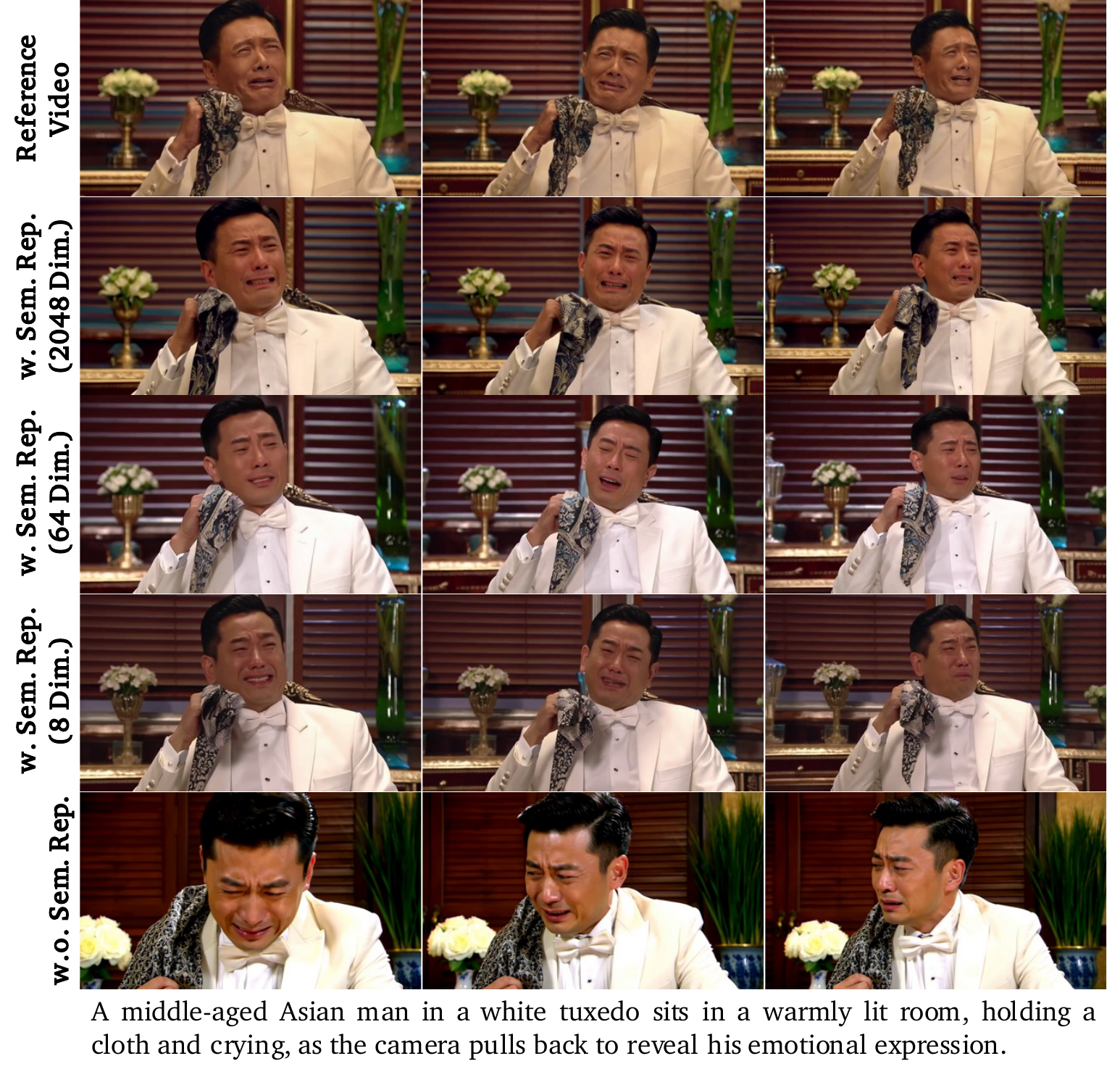}
\vspace{-0.6cm}
\caption{\textbf{Video generation conditioned on semantic features extracted from a reference video.} \textit{Row 1:} The reference video. \textit{Rows 2–4:} Reconstructions based on semantic representations (Sem. Rep.) with dimensions of 2048, 64, and 8, respectively. \textit{Row 5:} T2V generation results without semantic representations.}
    \vspace{-0.7cm}
    \label{fig_recons}
\end{figure}
\paragraph{Semantic Representation Compression for Effective Training.}
In our experiment, we empirically find that directly optimizing a pre-trained video diffusion model to fit high-dimensional semantic representations can result in slower convergence and inferior performance with fixed training steps (please refer to \cref{fig_mlp_ablation}). We hypothesize there are two-fold reasons. First, the high dimensionality of the semantic features leads to rich information, which may require longer convergence time during training. Second, the original semantic space is not conducive to sampling by a diffusion model. Therefore, we use a learnable MLP to compress the semantic space for effective training. The MLP serves two purposes: it reduces the dimensionality of the semantic representation and models the compressed feature space as a Gaussian distribution. The MLP outputs the mean and variance of this distribution, and we add the KL divergence objective as the regularizer, encouraging the learned compressed semantic space to resemble a Gaussian distribution. The sampled embedding $z_{sem}$ is then input into the diffusion model. This approach alleviates the fitting complexity for the semantic representation generation model, which will be introduced in \cref{sec:stage1}.
\vspace{-0.3cm}
\paragraph{In-Context Conditioning.}
The pipeline of the VAE latent generation stage is illustrated in \cref{fig:pipe}a. During training, we first feed the input video to the semantic encoder and the learnable MLP to get its compact semantic representation $z_{sem}$, which is then injected into the diffusion model via in-context conditioning \cite{recammaster}. Specifically, we concatenated noised VAE latents $z_{t}$ and compressed semantic representations $z_{sem}$ as the model's input, i.e., $z_{input} := [z_{t}, z_{sem}]$. 
To verify that the compressed semantic representation captures the video’s high-level semantics and effectively guides generation, we extract semantic features from a reference video and inject them into the VAE latent generator. The generated video, shown in \cref{fig_recons}, preserves the spatial layout and motion patterns of the reference video while differing in fine details. This demonstrates that the compressed semantic representations encode high-level information—such as structure and dynamics, while discarding low-level attributes like texture and color.
At the inference stage, $z_{sem}$ is generated by the semantic representation generation model in \cref{sec:stage1}. Similar to RAE \cite{rae}, we add noise to $z_{sem}$ to reduce the training-inference gap.

\subsection{Semantic Representation Generation}
\label{sec:stage1}
After training a VAE latent generator to synthesize VAE latents with the compressed semantic representations in \cref{sec:stage2} (illustrated in \cref{fig:pipe}a), we further learn the semantic representation distribution by another video diffusion model (illustrated in \cref{fig:pipe}b). In this stage, we freeze the visual encoder and the MLP, and fine-tune only the latent diffusion model. We observe a significant improvement in convergence speed after regularizing the semantic space with the learnable MLP. The results are summarized in \cref{fig_mlp_ablation} and \cref{tab:mlp_ablation}. Additionally, we ablate the design of using semantic encoders rather than VAE encoders in \cref{fig_vae_vit}, and we observe a significantly faster convergence speed compared to modeling the compressed VAE latents.

\definecolor{lightyellow}{HTML}{FFF2CC}
\definecolor{lightblue}{HTML}{DEEBF7}

\subsection{Extension to Long Video Generation}
\label{sec:longvideo}
\begin{figure}[t]\centering
\includegraphics[width=0.48\textwidth]{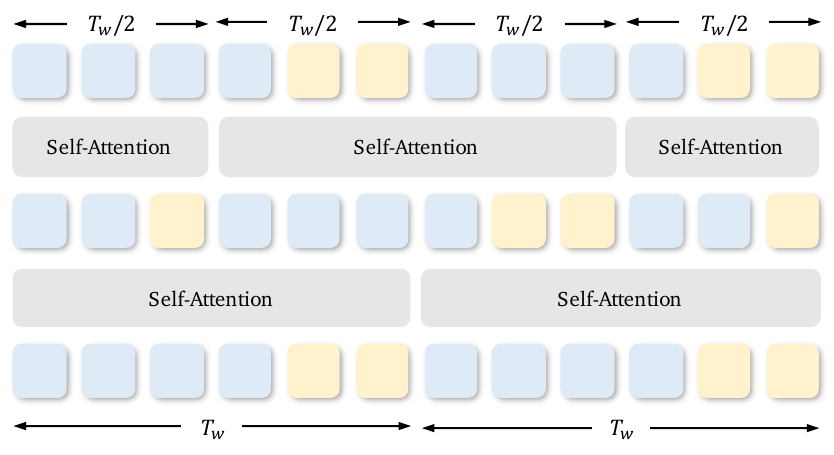}\vspace{-0.5em}
\caption{\textbf{Implementation of Swin-Attention.} When generating long videos, we apply full attention to model the semantic representations and use shifted-window attention \cite{swin} to map them into the VAE space. The \raisebox{0pt}[0pt][0pt]{\colorbox{lightblue!60}{blue}} squares indicate VAE latents, while the \raisebox{0pt}[0pt][0pt]{\colorbox{lightyellow!60}{yellow}} squares denote semantic representations.}
    \label{fig:swin}
\vspace{-1.0em}
\end{figure}

Previous diffusion-based video generation approaches \cite{kling, wan, seawead} often struggle to effectively scale to long video generation scenarios. This is because the computational cost of bi-directional attention increases quadratically with video length. Directly training the entire video (e.g., 1 minute) in the VAE latent space introduces unacceptable computational complexity. We propose to tackle this problem with SemanticGen. Our core insight is that, when generating long videos, we perform full-attention modeling only in the highly compressed semantic space to maintain consistency across scenes and characters in the video. When mapping to the VAE latent space, we use shifted window attention \cite{swin} to ensure that the computational cost does not grow quadratically with the number of frames. Since the semantic space naturally has a high compression ratio — in our implementation, the number of tokens is only 1/16 of the VAE tokens — the process of semantic representation generation introduces only a few additional computational costs. Meanwhile, the implementation of the shifted window attention in the VAE latent generation stage significantly reduces the model's computational cost compared to previous methods.
We illustrate the implementation of Swin attention in \cref{fig:swin}. To be specific, we interleave the VAE latent and their semantic representations, placing both types of tokens from a video of length $T_{w}$ into an attention window. The window then shifted by half a window size $T_{w}/2$ at odd layers.
\begin{figure*}[t]\centering
    \includegraphics[width=0.97\textwidth]{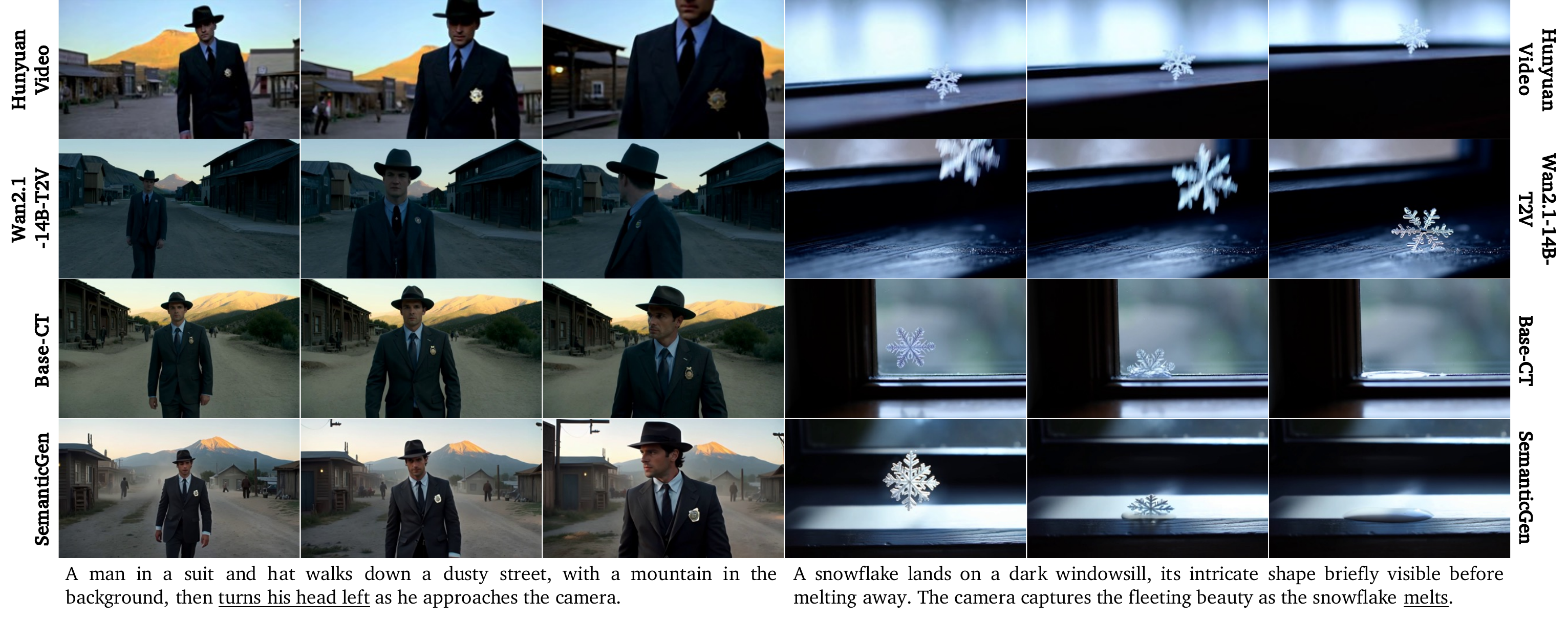}
    \vspace{-0.8em}
    \caption{\textbf{Comparison with state-of-the-art methods on short video generation.} It shows that SemanticGen generates high-quality videos that adhere to the text prompts and are comparable to strong baselines.}
    \label{fig_comparison_short}\vspace{-1.0em}
\end{figure*}

\begin{figure*}[t]\centering
    \includegraphics[width=0.94\textwidth]{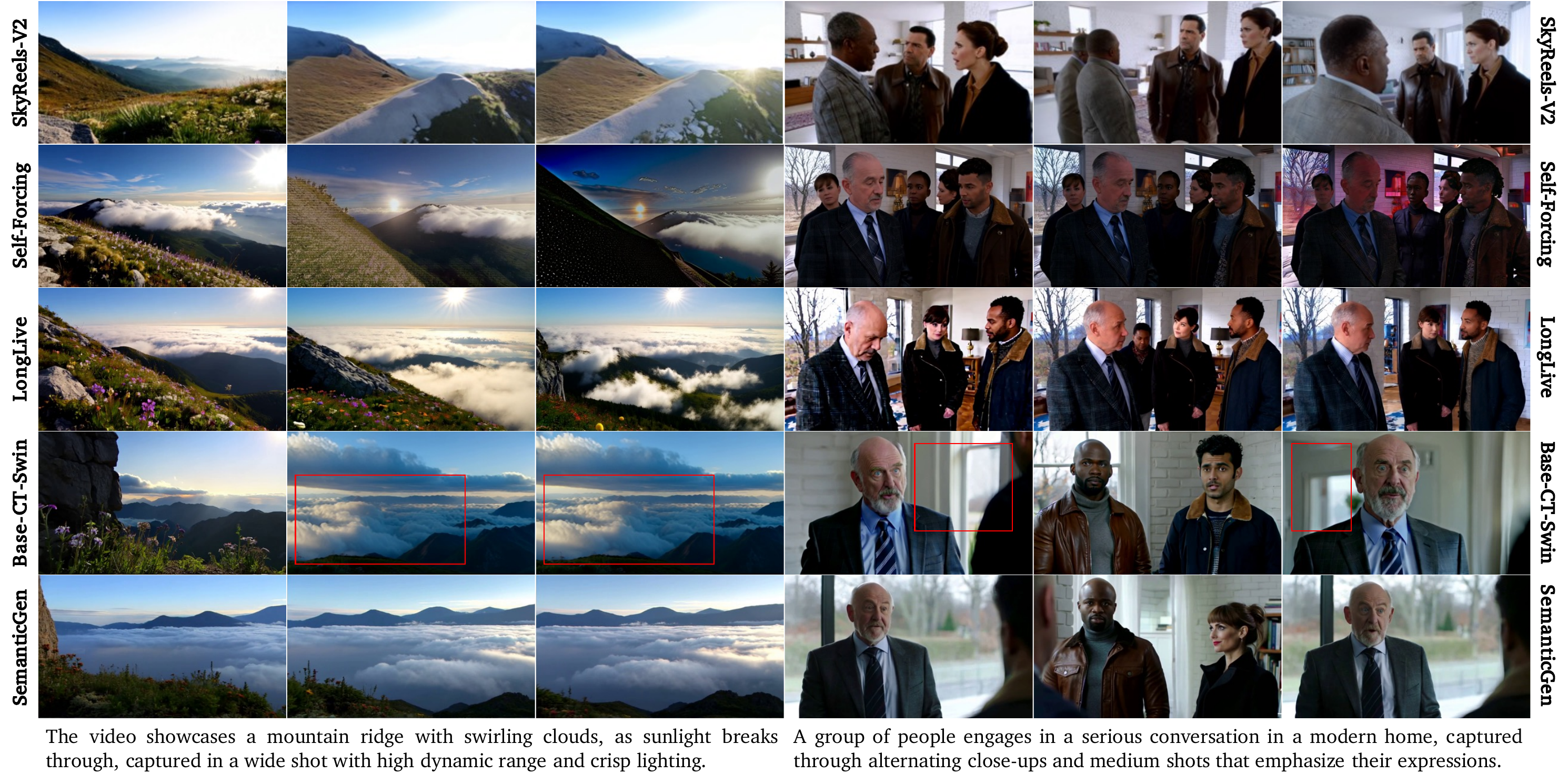}
    \vspace{-0.8em}
    \caption{\textbf{Comparison with state-of-the-art methods on long video generation.} It demonstrates that SemanticGen generates videos with long-term consistency and significantly alleviates the drifting issues.}
    \label{fig_comparison_long}\vspace{-1.0em}
\end{figure*}
\section{Experimental Results}
\label{sec:results}
\subsection{Experiment Settings}
\paragraph{Implementation Details.}
For short video generation, we train SemanticGen on an internal text-video pair dataset. For long video generation, we construct long video training data by splitting movie and TV show clips into 60-second segments and using an internal captioner to generate the corresponding text prompts. During training, we sample frames from the video at fps=24 as input to the VAE, and at fps=1.6 as input to the semantic encoder. We use the vision tower of Qwen2.5-VL-72B-Instruct \cite{qwen25vl} as the semantic encoder in our main experiments.

\begin{table*}[t]
	\begin{center}
            \vspace{-0.30cm}
		\caption{Quantitative comparison with state-of-the-art methods on short video generation.}
            \vspace{-0.30cm}
		\label{tab:short_video_gen}
		\setlength\tabcolsep{9.5pt}
		\begin{tabular}{lcccccc}
			\toprule
                Method & \makecell[c]{Subject\\Consistency} & \makecell[c]{Background\\Consistency} & \makecell[c]{Temporal\\Flickering} & \makecell[c]{Motion\\Smoothness} & \makecell[c]{Imaging\\Quality} & \makecell[c]{Aesthetic\\Quality} \\
                \midrule 
                Hunyuan-Video \cite{hunyuanvideo} & 91.11\% & 95.32\% & 97.49\% & 99.07\% & 64.23\% & 62.60\% \\
                Wan2.1-T2V-14B \cite{wan} & 97.23\% & \textbf{98.28\%} & 98.35\% & 99.08\% & \textbf{66.63\%} & \textbf{65.61\%} \\
                Base-CT & 96.17\% & 97.27\% & 98.07\% & 99.07\% & 65.77\% & 63.97\% \\
                SemanticGen & \textbf{97.79\%} & 97.68\% & \textbf{98.47\%} & \textbf{99.17\%} & 65.23\% & 64.60\% \\
			\bottomrule
		\end{tabular}
	\end{center}
        \vspace{-0.5cm}
\end{table*}
\begin{table*}[t]
	\begin{center}
		\caption{Quantitative comparison with state-of-the-art methods on long video generation.}
            \vspace{-0.30cm}
		\label{tab:long_video_gen}
		\setlength\tabcolsep{7.5pt}
		\begin{tabular}{lcccccccc}
			\toprule
                Method & \makecell[c]{Subject\\Consistency} & \makecell[c]{Background\\Consistency} & \makecell[c]{Temporal\\Flickering} & \makecell[c]{Motion\\Smoothness} & \makecell[c]{Imaging\\Quality} & \makecell[c]{Aesthetic\\Quality} & $\Delta^M_{drift}$\\
                \midrule 
                SkyReels-V2 \cite{skyreels-v2} & 93.13\% & 95.11\% & 98.41\% & 99.24\% & 66.00\% & 62.17\% & 9.00\% \\
                Self-Forcing \cite{self-forcing} & 90.41\% & 93.42\% & 98.51\% & 99.17\% & 70.23\% & 62.73\% & 12.39\% \\
                LongLive \cite{longlive} & 94.77\% & 95.90\% & 98.48\% & 99.21\% & 70.17\% & \textbf{64.73\%} & 4.08\% \\
                Base-CT-Swin & 94.01\% & 94.84\% & \textbf{98.64\%} & 99.32\% & 68.15\% & 61.66\% & 5.20\% \\
                SemanticGen & \textbf{95.07\%} & \textbf{96.70\%} & 98.31\% & \textbf{99.55\%} & \textbf{70.47\%} & 64.09\% & \textbf{3.58\%} \\
			\bottomrule
		\end{tabular}
	\end{center}
        \vspace{-0.5cm}
\end{table*}
\begin{figure*}[t]\centering
\includegraphics[width=0.99\textwidth]{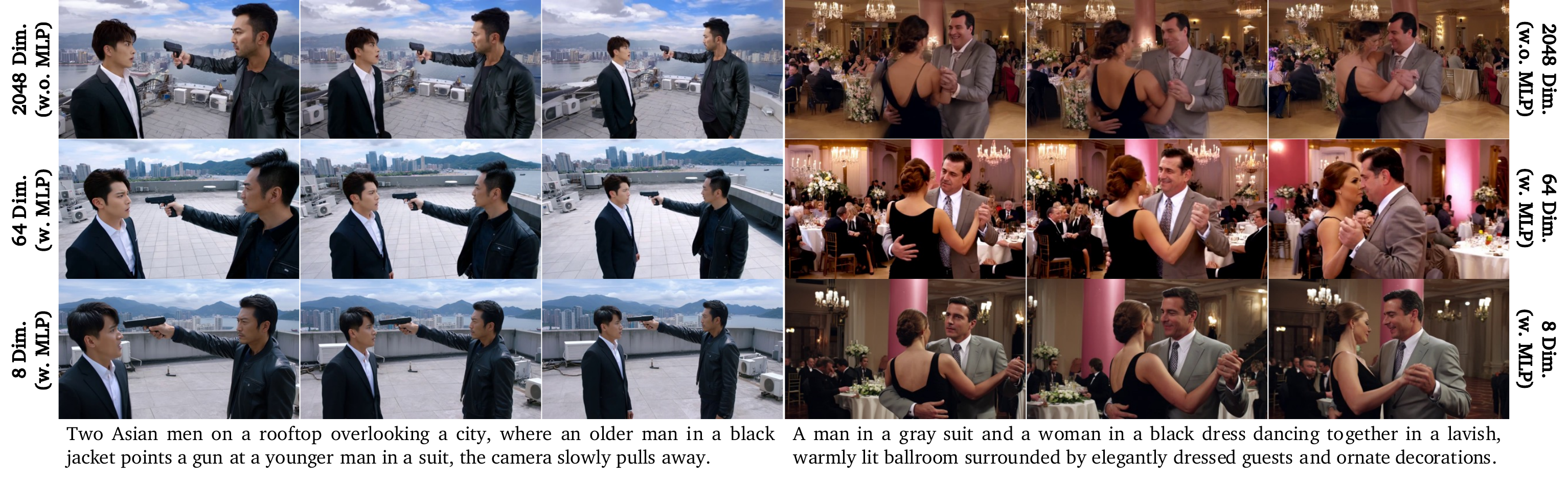}\vspace{-1em}
\caption{\textbf{Qualitative ablation on semantic space compression.} \textit{Row 1:} SemanticGen without compression; \textit{Row 2:} Compress the semantic space using an MLP with 64 output channels; \textit{Row 3:} Compress the semantic space using an MLP with 8 output channels.}
    \label{fig_mlp_ablation}
\vspace{-0.5em}
\end{figure*}

\begin{table*}[t]
	\begin{center}
		\caption{Quantitative ablation of semantic space compression with respect to different representation dimensionalities.}
            \vspace{-0.30cm}
		\label{tab:mlp_ablation}
		\setlength\tabcolsep{8.0pt}
		\begin{tabular}{lccccccc}
			\toprule
                Method & \makecell[c]{Subject\\Consistency} & \makecell[c]{Background\\Consistency} & \makecell[c]{Temporal\\Flickering} & \makecell[c]{Motion\\Smoothness} & \makecell[c]{Imaging\\Quality} & \makecell[c]{Aesthetic\\Quality}\\
                \midrule 
                w.o. compression (dim=2048) & 96.29\% & 96.54\% & 96.39\% & 99.31\% & 67.42\% & 58.88\% \\
                w. compression (dim=64) & 97.36\% & 96.85\% & 98.23\% & 98.34\% & 68.16\% & 60.62\% \\
                w. compression (dim=8) & \textbf{97.49\%} & \textbf{97.34\%} & \textbf{98.27\%} & \textbf{99.38\%} & \textbf{68.43\%} & \textbf{60.95\%} \\
			\bottomrule
		\end{tabular}
	\end{center}
        \vspace{-0.5cm}
\end{table*}

\vspace{-0.1cm}
\paragraph{Evaluation Set and Metrics.}
We evaluate short- and long-form video generation using the standard VBench \cite{vbench} and VBench-Long \cite{vbench++} benchmarks. We extend their official prompt sets and apply them to SemanticGen and all baselines. Video quality is assessed using the standard VBench metrics. For long video generation, we additionally measure quality drift using $\Delta^M_{\text{drift}}$, proposed in FramePack \cite{framepack}, defined as the absolute value of the difference in metric values between the initial and final segments of a video.
\subsection{Comparison with State-of-the-Art Methods}

\begin{figure}[t]\centering
\includegraphics[width=0.46\textwidth]{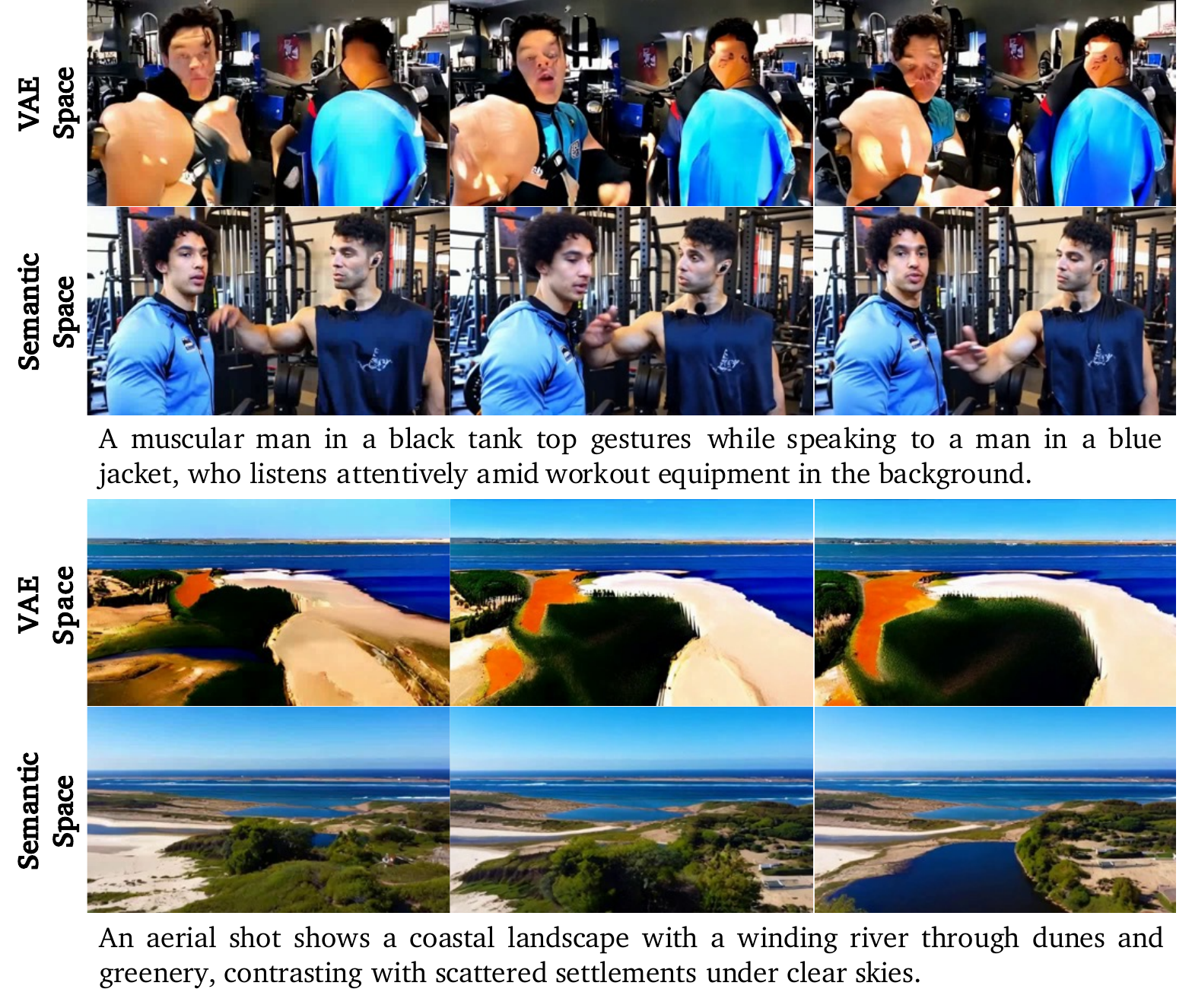}
\vspace{-0.2cm}
\caption{\textbf{Ablation on the representation space.} We visualize the generation results of learning on the semantic space and the compressed VAE latent space with the same training steps.}
    \vspace{-0.5cm}
    \label{fig_vae_vit}
\end{figure}

\paragraph{Baselines.}  
We compared the proposed SemanticGen with state-of-the-art T2V methods in \cref{tab:short_video_gen} and \cref{tab:long_video_gen}. For short video generation, we use Wan2.1-T2V-14B \cite{wan}, and HunyuanVideo \cite{hunyuanvideo} as baselines. For long video generation, we use open-source models SkyReels-V2 \cite{skyreels-v2}, Self-Forcing \cite{self-forcing}, and LongLive \cite{longlive} as baselines.
It is important to note that existing video generation studies typically use different base models, training data, and training steps, making fair comparisons challenging. To provide a reliable assessment of our proposed paradigm, we include additional baselines that continue training the base model using the standard diffusion loss without semantic modeling, while keeping the data and the number of training steps identical. These comparisons are included as important baselines in \cref{tab:short_video_gen} and \cref{tab:long_video_gen}, denoted as Base-CT and Base-Swin-CT.
\vspace{-0.2cm}
\paragraph{Qualitative Results.}
We present synthesized examples of SemanticGen in \cref{fig_1}. Please refer to the \href{https://jianhongbai.github.io/SemanticGen/}{project page} for video results. SemanticGen demonstrates the ability to: 1) generate high-quality videos adhering to the text prompts; 2) generate long videos with
long-term consistency, and significantly alleviate drifting issues.
We compare SemanticGen with state-of-the-art methods on short video generation and long video generation in Fig. \ref{fig_comparison_short} and Fig. \ref{fig_comparison_long}, respectively. For short video generation, SemanticGen surpasses baseline methods in text-following accuracy. For example, the baselines fail to generate the man turning his head to the left or the melting process of the snowflake. Compared with continuing to train the base model using the diffusion-based framework (denoted as Base-CT in \cref{fig_comparison_short}), our method achieves comparable performance. For long video generation, SemanticGen achieves better long-term consistency and significantly alleviates drifting issues. We observe that baselines may exhibit severe color shifts or inconsistencies across frames. Similar phenomena appear when continuing to train the base text-to-video generation model with Swin attention without global semantic modeling (denoted as Base-CT-Swin in \cref{fig_comparison_long}), where we observe inconsistent backgrounds across generated frames and more artifacts, highlighting the importance of performing global planning in the high-level semantic space.
\vspace{-0.2cm}
\paragraph{Quantitative Results.}
We quantitatively evaluate SemanticGen against baselines using automatic metrics, with results summarized in Tables \ref{tab:short_video_gen} and \ref{tab:long_video_gen}. We adopt the VBench metrics to assess visual quality. For short-video generation, SemanticGen achieves performance comparable to state-of-the-art T2V models and the continued-training baseline of our base model. For long-video generation, SemanticGen substantially outperforms all baselines in terms of video consistency and temporal stability, benefiting from using full attention to model high-level semantic features that enhance long-term coherence. We also employ the drifting-measurement metric $\Delta^M_{drift}$ introduced in \cite{framepack} to quantify quality degradation over time. $\Delta^M_{drift}$ is defined as the difference in metric values between the first and last 15\% segments of a video. SemanticGen also surpasses baselines on this metric. Meanwhile, SemanticGen learns in a data-driven manner to generate coherent multi-shot videos.
\subsection{More Analysis and Ablation Studies}
\paragraph{The Effectiveness of Semantic Space Compression.}
In \cref{sec:stage2}, we propose to compress the semantic representation space using a lightweight MLP for efficient training. The effectiveness of this approach is validated through qualitative and quantitative results in \cref{fig_mlp_ablation} and \cref{tab:mlp_ablation}, respectively. Specifically, we use the vision tower of Qwen2.5VL-3B-Instruct \cite{qwen25vl} as the semantic encoder, where the vanilla semantic representation has a dimension of 2048. We first train three VAE latent generators (illustrated in \cref{fig:pipe}b) using: (1) no MLP, (2) an MLP with 64 output channels, and (3) an MLP with 8 output channels, each for 10K steps. Based on these models, we further train three corresponding semantic generation models (illustrated in \cref{fig:pipe}a) for 50K steps. During inference, we first use the semantic generator to produce the video semantic representation, which is then used as a condition input to the VAE latent generation model to map it into the VAE space.
As shown in \cref{fig_mlp_ablation}, we observe that the visual quality of the generated videos improves as the dimensionality decreases, exhibiting fewer broken frames and artifacts. We further quantitatively evaluate the video quality on VBench metrics with 47 text prompts in \cref{tab:mlp_ablation}, which also confirms this trend. This indicates that compressing the pre-trained semantic representation space to a lower dimension accelerates the convergence of the semantic generator.
\paragraph{SemanticGen Achieve Faster Convergence Speed.}

In this paper, we propose to first learn compact semantic representations and then map them into the VAE latent space. A natural question arises: Does leveraging semantic representations truly benefit video generation? In other words, what happens if we adopt the same two-stage pipeline but learn compact VAE latents instead of semantic representations \cite{tokensgen}?
To investigate this, we keep the SemanticGen framework unchanged except for replacing the semantic encoder with a VAE encoder, training a generator to model compressed VAE latents rather than semantic features. Both the semantic generator and the VAE latent generator are trained from scratch for 10K steps, and the results are shown in \cref{fig_vae_vit}. We observe that modeling in the VAE space leads to significantly slower convergence, as the generated results only contain coarse color patches. In contrast, the model trained in the semantic space is already able to produce reasonable videos under the same number of training steps. This demonstrates that the proposed SemanticGen framework effectively accelerates the convergence of diffusion-based video generation models.

\section{Conclusion and Limitations}
\vspace{0.05cm}
In this paper, we propose SemanticGen, a video generation framework that synthesizes videos in a compact semantic space. The key idea is to first generate high-level semantic representations for global planning and then refine them with high-frequency details. SemanticGen follows a two-stage pipeline: it first produces semantic video features that define the global layout, and then generates VAE latents conditioned on these features to produce the final video. We observe faster convergence in the semantic space than in the VAE latent space, and the approach scales efficiently to long video generation.
Despite these advantages, several limitations remain. Long video generation struggles to maintain consistency in textures, as semantic features cannot fully preserve fine-grained details. In addition, SemanticGen inherits constraints from its semantic encoders. For instance, sampling at low fps leads to the loss of high-frequency temporal information, as shown in the appendix.
{
    \small
    \bibliographystyle{ieeenat_fullname}
    \bibliography{main}
}

\clearpage
\setcounter{page}{1}
\maketitlesupplementary

\begin{appendices}
\begin{figure*}[t]\centering
\includegraphics[width=0.98\textwidth]{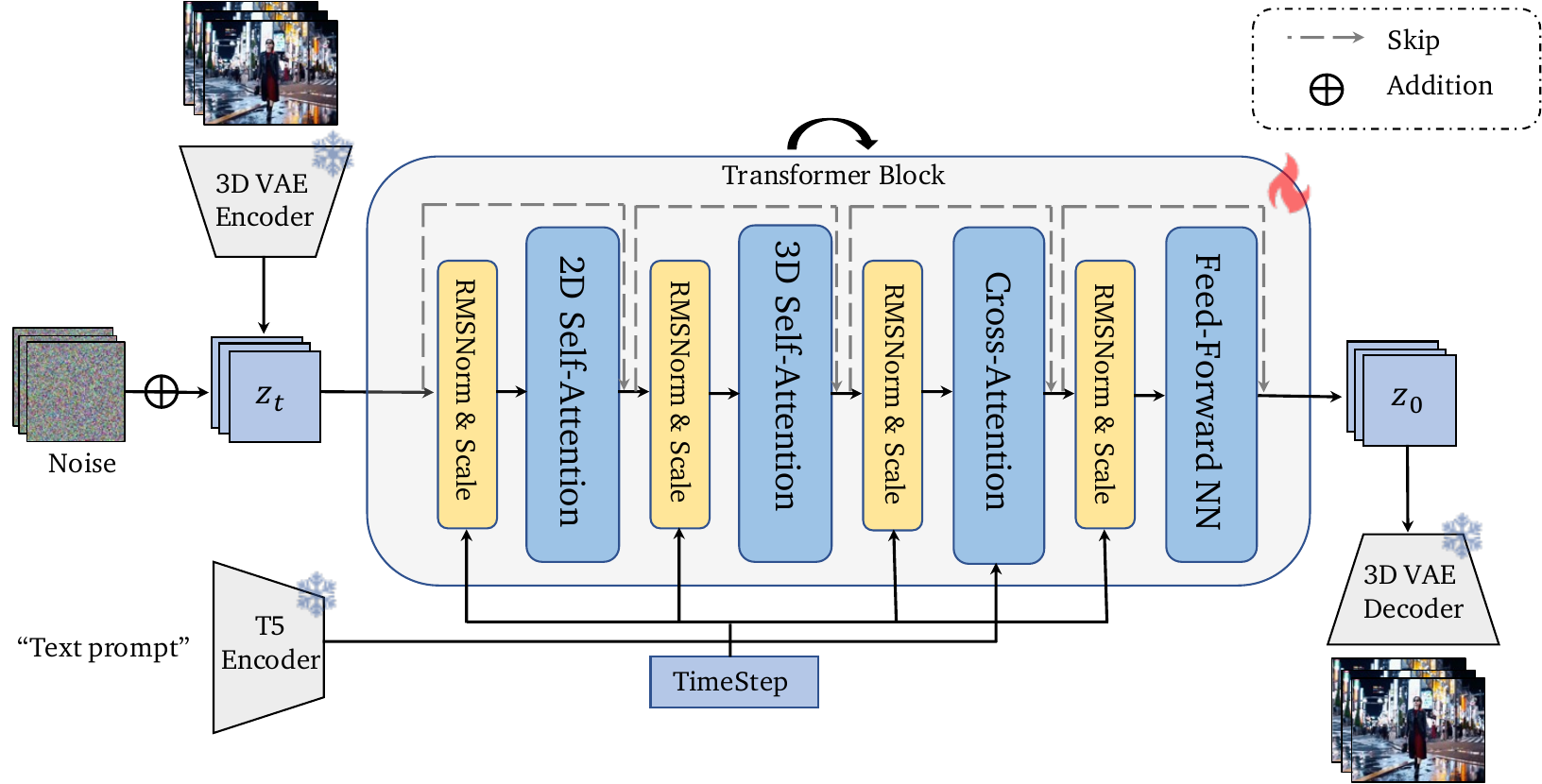}
\caption{\text{Overview of the base text-to-video generation model.}}
    \label{fig_basemodel}
\end{figure*}
\section{Introduction of the Base Text-to-Video Generation Model}
We use a transformer-based latent diffusion model \citep{dit} as the base T2V generation model, as illustrated in Fig. \ref{fig_basemodel}. We employ a 3D-VAE to transform videos from the pixel space to a latent space, upon which we construct a transformer-based video diffusion model. We use 3D self-attention, enabling the model to effectively perceive and process spatiotemporal tokens, thereby achieving a high-quality and coherent video generation model. Specifically, before each attention or feed-forward network (FFN) module, we map the timestep to a scale, thereby applying RMSNorm to the spatiotemporal tokens.

\section{More Results}
\label{sec:appendix_results}

\subsection{More Results of SemanticGen}
\begin{figure*}[t]\centering
\includegraphics[width=1\textwidth]{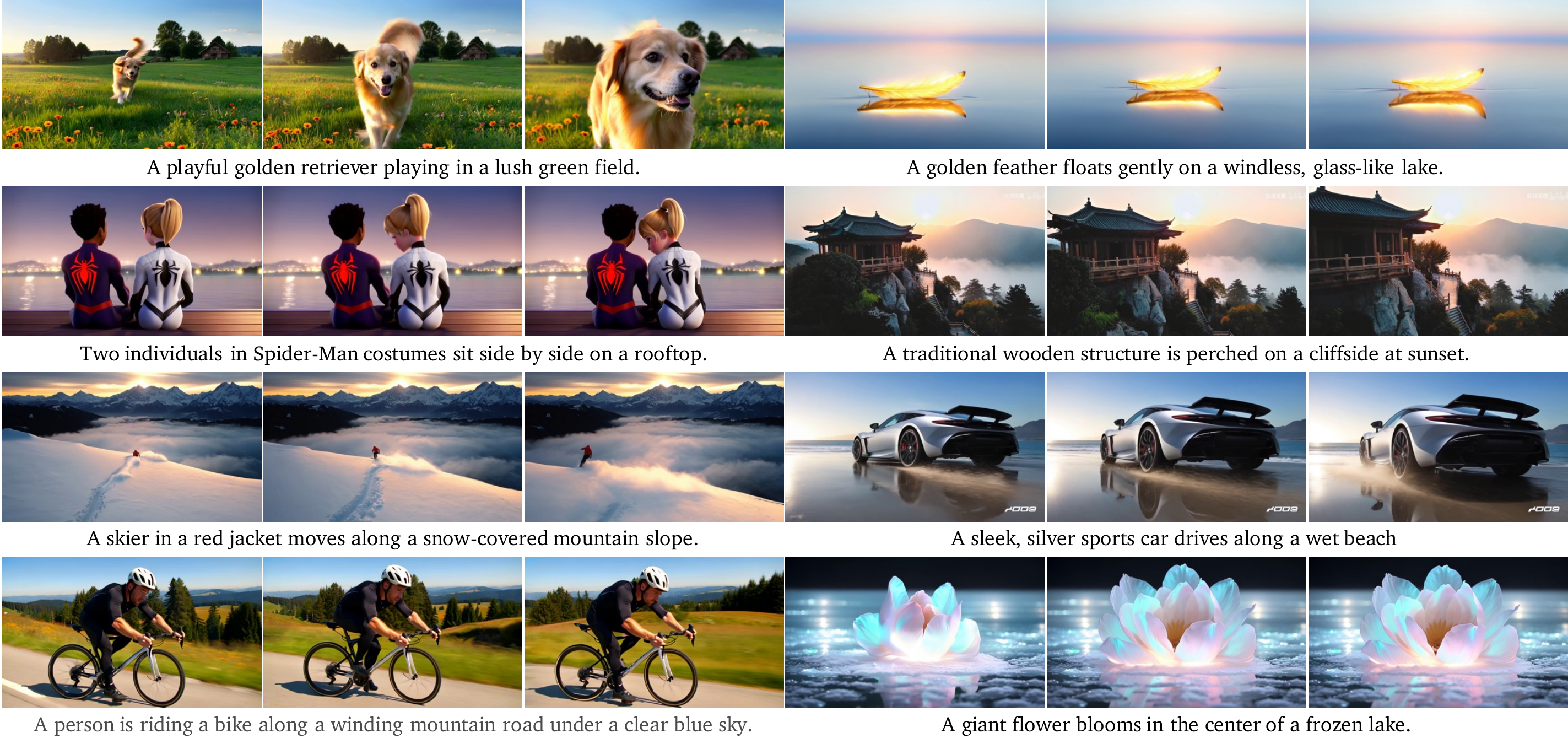}
\caption{More synthesized results of SemanticGen.}
    \label{fig_appendix_ours1}
\end{figure*}

\begin{figure*}[t]\centering
\includegraphics[width=1\textwidth]{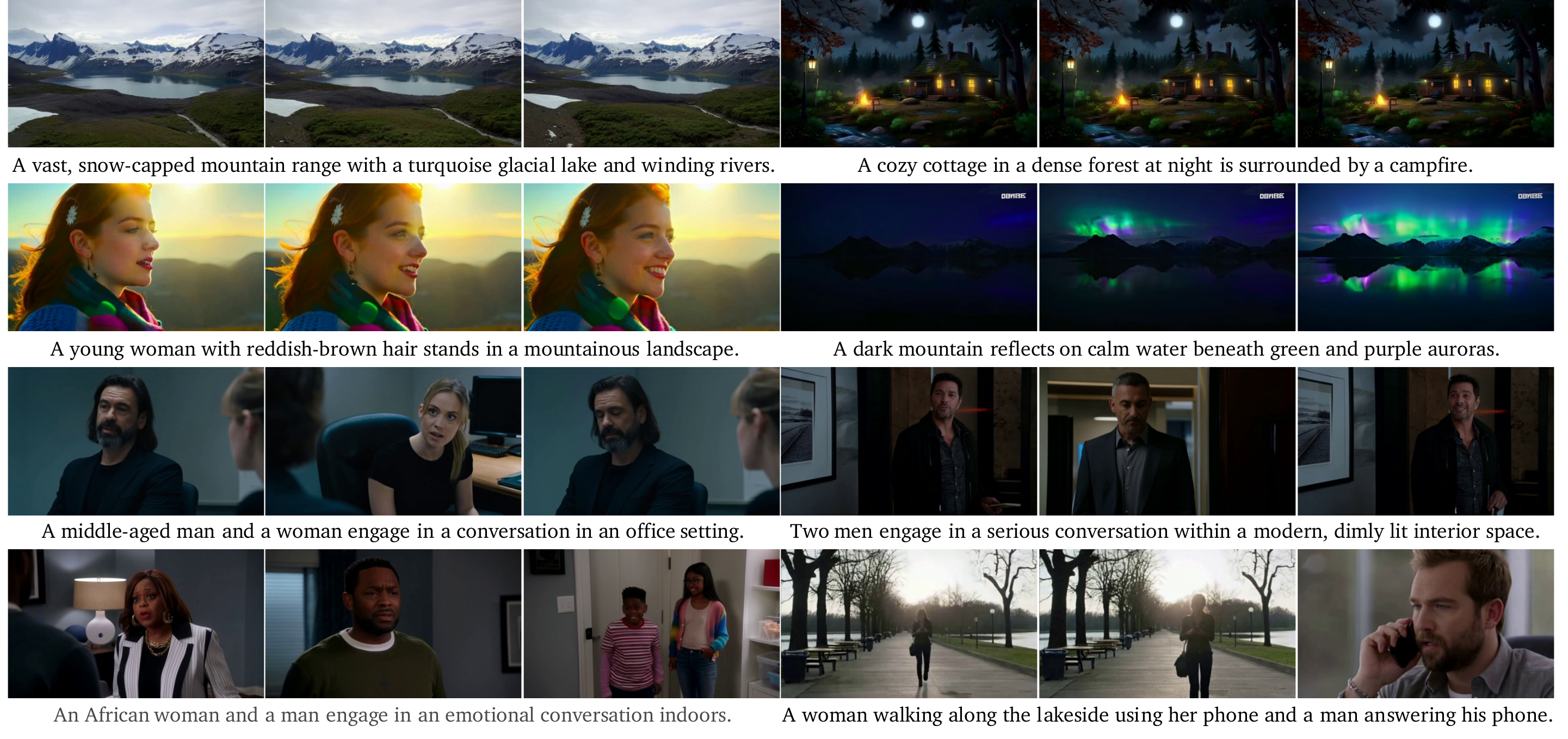}
\caption{More synthesized results of SemanticGen.}
    \label{fig_appendix_ours2}
\end{figure*}
More synthesized results of SemanticGen are presented in \cref{fig_appendix_ours1} and \cref{fig_appendix_ours2}. SemanticGen demonstrates the ability to: 1) generate high-quality videos adhering to the text prompts; 2) generate long videos with
long-term consistency, and significantly alleviate drifting issues. Please refer to the project page for video results.
\subsection{Comparison with Additional Baselines}
\begin{figure*}[t]\centering
\includegraphics[width=1\textwidth]{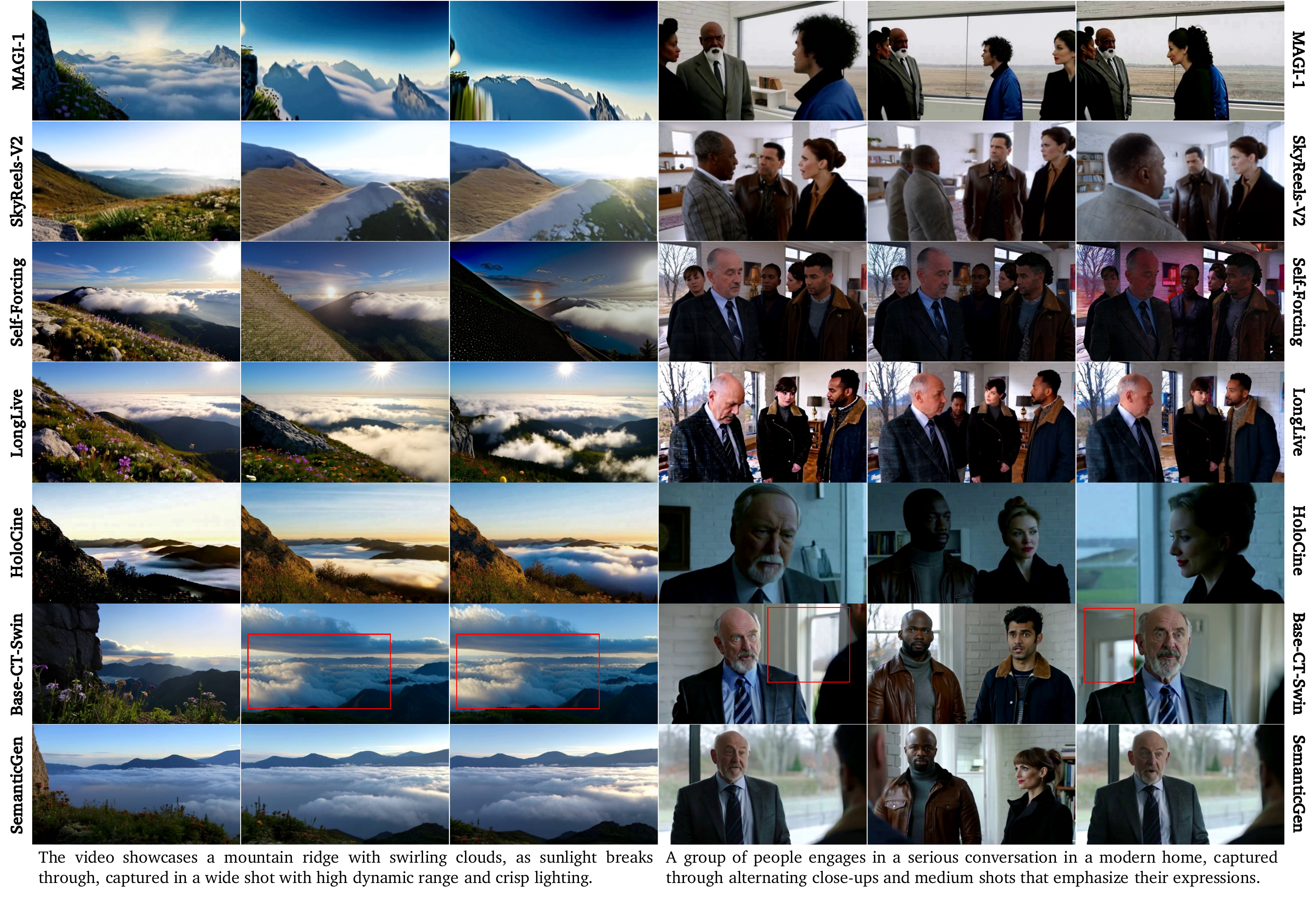}
\caption{Comparison with additional baselines.}
    \label{fig_appendix_comparison}
\end{figure*}
We include qualitative comparisons on long video generation with additional baselines in \cref{fig_appendix_comparison}. We use open-source models MAGI-1 \cite{magi-1}, SkyReels-V2 \cite{skyreels-v2}, Self-Forcing \cite{self-forcing}, LongLive \cite{longlive}, and HoloCine \cite{holocine} as baselines. To provide a reliable assessment of our proposed paradigm, we include additional baselines that continue training the base model using the standard diffusion loss without semantic modeling, while keeping the data and the number of training steps identical. These comparisons are included as important baselines in \cref{fig_appendix_comparison}, denoted as Base-CT and Base-Swin-CT. SemanticGen achieves better long-term consistency and significantly alleviates drifting issues. We observe that baselines may exhibit severe color shifts or inconsistencies across frames. Similar phenomena appear when continuing to train the base text-to-video generation model with Swin attention without global semantic modeling (denoted as Base-CT-Swin in \cref{fig_appendix_comparison}), where we observe inconsistent backgrounds across generated frames and more artifacts, highlighting the importance of performing global planning in the high-level semantic space.

\subsection{Failure Cases Visualization}

\begin{figure*}[!t]\centering
\includegraphics[width=1\textwidth]{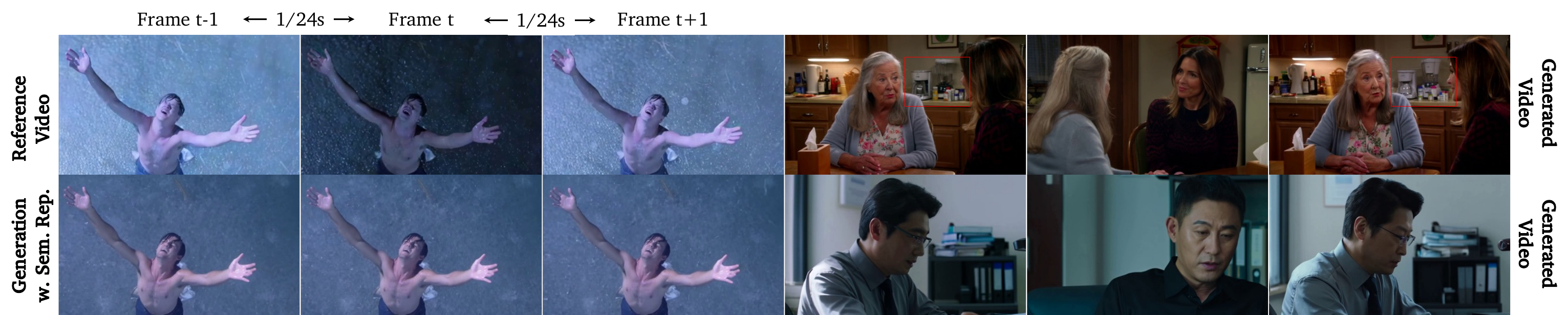}
\vspace{-0.2cm}
\caption{Visualization of failure cases.}
    \label{fig_failure}
    \vspace{-0.4cm}
\end{figure*}

We present the failure cases in \cref{fig_failure}. Since our model utilizes pre-trained video understanding tokenizers as semantic encoders, SemanticGen inherits constraints from its understanding tokenizers. For instance, sampling at low fps as the input to the understanding tokenizers leads to the loss of high-frequency temporal information. To illustrate this phenomenon, we input a reference video into the semantic encoder to extract semantic representations, and then feed these representations into the VAE latent generator to produce a video. As shown in the first two rows of \cref{fig_failure}, when we sample at fps=1.6 as the input to the semantic encoder, it fails to capture the temporal variation within 1/24 of a second (e.g., the sky changing from bright to dark and back due to lightning), resulting in the generated video lacking flicker. We anticipate the development of video understanding tokenizers that simultaneously achieve high temporal compression and high sampling rates, which could further enhance SemanticGen’s performance. Additionally, we observed that for long video generation, fine-grained details (such as textures or small objects) may not be consistently preserved, as semantic features cannot fully capture these details, as shown in the last row of \cref{fig_failure}.

\section{Future Work}
\label{sec:appendix_future_work}

\paragraph{Systematic Analysis of Different Semantic Encoders} In this paper, we propose first modeling compact semantic representations and then mapping them into the VAE latent space. We leverage the vision tower of Qwen-2.5-VL \cite{qwen25vl} as the semantic encoder to demonstrate the effectiveness of SemanticGen. A systematic analysis of using different semantic encoders \cite{v-jepa2, videomae2, 4ds} is valuable. Specifically, it is important to explore whether the generated performance varies when using semantic encoders trained with different paradigms (e.g., visual-text alignment, self-supervised learning, etc.).

\paragraph{Towards More Informative Video Semantic Encoders} Pre-trained video semantic encoders play a vital role in SemanticGen, as generation first occurs in the semantic space. Therefore, a more powerful semantic encoder could lead to better generation performance. For example, we need a tokenizer that not only achieves high temporal compression but also samples the original video at a high frame rate, which would better facilitate modeling high-frequency temporal information.

\end{appendices}

\end{document}